\documentclass[sigconf,nonacm]{acmart}
\usepackage[ruled,linesnumbered]{algorithm2e}
\usepackage{multirow}
\usepackage[normalem]{ulem}
\usepackage{url}
\usepackage{hyperref}
\useunder{\uline}{\ul}{}

\AtBeginDocument{%
  }

\setcopyright{acmlicensed}
\copyrightyear{2018}
\acmYear{2018}
\acmDOI{XXXXXXX.XXXXXXX}
\acmConference[Conference acronym 'XX]{Make sure to enter the correct
  conference title from your rights confirmation email}{June 03--05,
  2018}{Woodstock, NY}
\acmISBN{978-1-4503-XXXX-X/2018/06}

\begin{document}

\title{MaGNet: A Mamba Dual-Hypergraph Network for Stock Prediction via Temporal-Causal and Global Relational Learning}

\author{Peilin Tan}
\affiliation{%
  \institution{University of California, San Diego}
  \city{La Jolla}
    \state{CA}
  \country{USA}}
\email{p9tan@ucsd.edu}

\author{Chuanqi Shi}
\affiliation{%
  \institution{University of California, San Diego}
  \city{La Jolla}
    \state{CA}
  \country{USA}}
\email{chs028@ucsd.edu}

\author{Dian Tu}
\affiliation{%
  \institution{University of California, San Diego}
  \city{La Jolla}
    \state{CA}
  \country{USA}}
\email{ditu@ucsd.edu}

\author{Liang Xie}
\authornote{Corresponding author.}
\affiliation{%
  \institution{Wuhan University of Technology}
  \city{Wuhan, Hubei}
  \country{China}
}
\email{whutxl@hotmail.com}

\begin{abstract}
Stock trend prediction is crucial for profitable trading strategies and portfolio management yet remains challenging due to market volatility, complex temporal dynamics and multifaceted inter-stock relationships. Existing methods struggle to effectively capture temporal dependencies and dynamic inter-stock interactions, often neglecting cross-sectional market influences, relying on static correlations, employing uniform treatments of nodes and edges, and conflating diverse relationships. This work introduces MaGNet, a novel \textbf{Ma}mba dual-hyper\textbf{G}raph \textbf{Net}work for stock prediction, integrating three key innovations: (1) a MAGE block, which leverages bidirectional Mamba with adaptive gating mechanisms for contextual temporal modeling and integrates a sparse Mixture-of-Experts layer to enable dynamic adaptation to diverse market conditions, alongside multi-head attention for capturing global dependencies; (2) Feature-wise and Stock-wise 2D Spatiotemporal Attention modules enable precise fusion of multivariate features and cross-stock dependencies, effectively enhancing informativeness while preserving intrinsic data structures, bridging temporal modeling with relational reasoning; and (3) a dual hypergraph framework consisting of the Temporal-Causal Hypergraph (TCH) that captures fine-grained causal dependencies with temporal constraints, and Global Probabilistic Hypergraph (GPH) that models market-wide patterns through soft hyperedge assignments and Jensen-Shannon Divergence weighting mechanism, jointly disentangling localized temporal influences from instantaneous global structures for multi-scale relational learning. Extensive experiments on six major stock indices demonstrate MaGNet outperforms state-of-the-art methods in both superior predictive performance and exceptional investment returns with robust risk management capabilities. \textbf{Datasets, source code, and model weights are available at our GitHub repository: https://github.com/PeilinTime/MaGNet}.
\end{abstract}



\maketitle

\section{Introduction}
The stock market serves as a crucial component of the global financial system and a primary investment avenue. Stock trend prediction, which forecasts future price movements to guide investment decisions and risk management, has garnered substantial attention. However, prediction remains inherently challenging due to high volatility, non-stationary behavior and complex influencing factors including macroeconomic conditions, company performance, and inter-stock relationships.

Stock price prediction has evolved from traditional methods (SVM, ARIMA \cite{adebiyi2014comparison}) to deep learning approaches. While CNNs \cite{CNN2}, RNNs (LSTMs \cite{LSTM2}), and Transformers \cite{attention2} improved capturing complex market behaviors, they face challenges with long-range dependencies and computational complexity. State Space Models like Mamba \cite{gu2023mamba} recently emerged as efficient alternatives, achieving near-linear complexity through selective scan mechanisms.

Stock correlation modeling progressed from static, predefined connections to dynamic representations. Graph Neural Networks (GNNs) \cite{chen2024automatic} model stocks as nodes with dynamic edges but only capture pairwise relationships. Since stock movements often involve higher-order group dynamics through shared industry membership or ownership, Hypergraph Neural Networks (HGNNs) \cite{HGNN2025AAAI} were introduced to connect multiple nodes simultaneously via hyperedges, though challenges remain in appropriately weighting neighbors and hyperedges.

Despite these advances, critical limitations persist in current approaches that need addressing: (1) For time series modeling, while Mamba offers efficient linear complexity through selective state space mechanisms, it lacks contextual modeling with sophisticated temporal fusion, adaptation to diverse market regimes, and global dependency capture—all crucial for financial markets. Moreover, existing temporal models treat each stock's time series independently, failing to preserve cross-sectional information embedded in the feature space that captures critical inter-stock dynamics \cite{li2019individualized}; (2) For relational modeling, while HGNNs advance beyond pairwise connections, they suffer from uniform treatment of nodes within hyperedges despite varying influence levels, and inability to distinguish between different types of relationships (e.g., causal vs. instantaneous, local vs. global). These limitations result in models that cannot fully capture the complex, dynamic and multi-scale nature of financial markets, necessitating a more sophisticated architecture.

To address these challenges, we propose a novel architecture combining advanced temporal modeling with dynamic relational learning. Our contributions are threefold:
\begin{itemize}
\item \textbf{MAGE Block with 2D Attention}: To fully capture temporal dynamics, we design the \textbf{MAGE} (\textbf{M}amba-\textbf{A}ttention-\textbf{G}ating-\textbf{E}xperts) block by enhancing bidirectional Mamba with adaptive gating to capture the full temporal context. We integrate sparse MoE for market regime adaptation and multi-head attention for global dependencies. We further apply feature-wise 2D spatiotemporal attention to enhance stock features by capturing dynamic interactions across features and stocks over time. This design jointly captures temporal dynamics and spatial patterns.
\item \textbf{Dual Hypergraph Framework}: 
To model dynamic and high-order relations among stocks, we introduce two complementary hypergraphs. The Temporal-Causal Hypergraph (TCH) captures fine-grained, localized relations across stocks and time. The Global Probabilistic Hypergraph (GPH) encodes broader market structures through probabilistic hyperedges, allowing stocks to have varying membership degrees across multiple groups rather than uniform treatment. This dual design disentangles local temporal-causal signals from global market patterns, enabling expressive and flexible relational learning.
\item \textbf{State-of-the-Art Performance}: Extensive experiments on six major stock indices demonstrate that MaGNet achieves superior predictive accuracy (up to 54.9\% on CSI 300) and exceptional investment returns with robust risk-adjusted performance, including Sharpe ratios exceeding 1.0 on multiple markets and annual returns up to 22.6\%.
\end{itemize}

\section{Related Work}
\subsection{Time Series Models for Stock Prediction}
Deep learning revolutionized stock prediction by capturing complex temporal patterns. RNNs and variants (LSTM, GRU) became widely adopted for modeling sequential dependencies \cite{shih2019temporal}. CNN-based approaches treated historical data as feature maps, with architectures like dilated CNNs capturing multi-scale patterns \cite{tsai2019finenet}. Attention mechanisms enhanced these models by focusing on relevant historical states \cite{qin2017dual}, culminating in transformer architectures that achieved remarkable results through self-attention \cite{ding2020hierarchical}.

Recently, State Space Models (SSMs) emerged as efficient alternatives, combining CNN-like parallel training with RNN-like fast inference. Mamba \cite{gu2023mamba}, a selective SSM with time-varying parameters and hardware-aware algorithms, has shown particular promise. Extensions include S-Mamba \cite{wang2025mamba} for multivariate series and bidirectional variants like MambaMixer \cite{behrouz2024mambamixerefficientselectivestate} for enhanced contextual modeling. However, sequential models often treat stocks independently, ignoring crucial inter-stock relationships.

\subsection{Graph and Hypergraph Methods for Inter-Stock Relationships}
Recognizing stock interdependencies, graph-based methods explicitly model relationships through various connections: shareholding, industry sectors \cite{HGNN4}, supply chains, and price correlations \cite{mehrabian2025mamba}. Graph Neural Networks (GNNs) learn representations over graphs through neighborhood aggregation, with GCNs combined with temporal models \cite{feng2019temporal} capturing evolving relationships and GATs \cite{velivckovic2017graph,HGNN3} using attention for refined message passing.

However, GNNs face limitations: reliance on predefined relationships that may be incomplete or noisy, and inability to model higher-order group interactions \cite{gao2021graph,attention1}. Hypergraph methods \cite{HGNN5,cui2023temporal,gao2022hgnn} address this by using hyperedges connecting multiple nodes simultaneously, naturally representing group-wise dependencies like industry sectors or shared fund ownership. Frameworks include HyperGCN \cite{Hypergcn} and STHGCN \cite{sawhney2020spatiotemporal} for spatiotemporal modeling. Despite these advances, many hypergraph models struggle with dynamic temporal features and uniform node treatment.

\section{Methodology}

\subsection{Problem Definition}

We address the task of predicting next-day stock price movement, framed as a binary classification problem. Let $\mathcal{S} = \{s_1, s_2, \ldots, s_N\}$ represent a set of $N$ stocks in the market. For each stock $s_i \in \mathcal{S}$, we consider its historical trading data over a lookback window of $T$ days, with $F$ financial indicators recorded daily, resulting in a feature matrix $\mathbf{X}_i \in \mathbb{R}^{T \times F}$ for stock $s_i$. The complete input data across all stocks can be represented as $\mathbf{X} = [\mathbf{X}_1, \mathbf{X}_2, \ldots, \mathbf{X}_N] \in \mathbb{R}^{N \times T \times F}$. The prediction target is the direction of all stocks' closing price movement on the following day, encoded as a binary label:
\begin{equation}
y_i = 
\begin{cases}
1, & \text{if } p_i^{(t+1)} > p_i^{(t)} \\
0, & \text{otherwise}
\end{cases}, \forall i \in \{1, 2, \ldots, N\},
\end{equation}
where $p_i^{(t)}$ denotes the closing price of stock $s_i$ on day $t$.

The objective is to learn a predictive function $f(\cdot; \boldsymbol{\Theta})$ parameterized by $\boldsymbol{\Theta}$ that maps $\mathbf{X}$ to the predicted probabilities of upward price movement for all stocks: $\hat{\mathbf{Y}} = f(\mathbf{X}; \boldsymbol{\Theta}), \enspace \hat{\mathbf{Y}} \in [0, 1]^N$.

\begin{figure*}[t]
\centering
\includegraphics[width=1\textwidth]{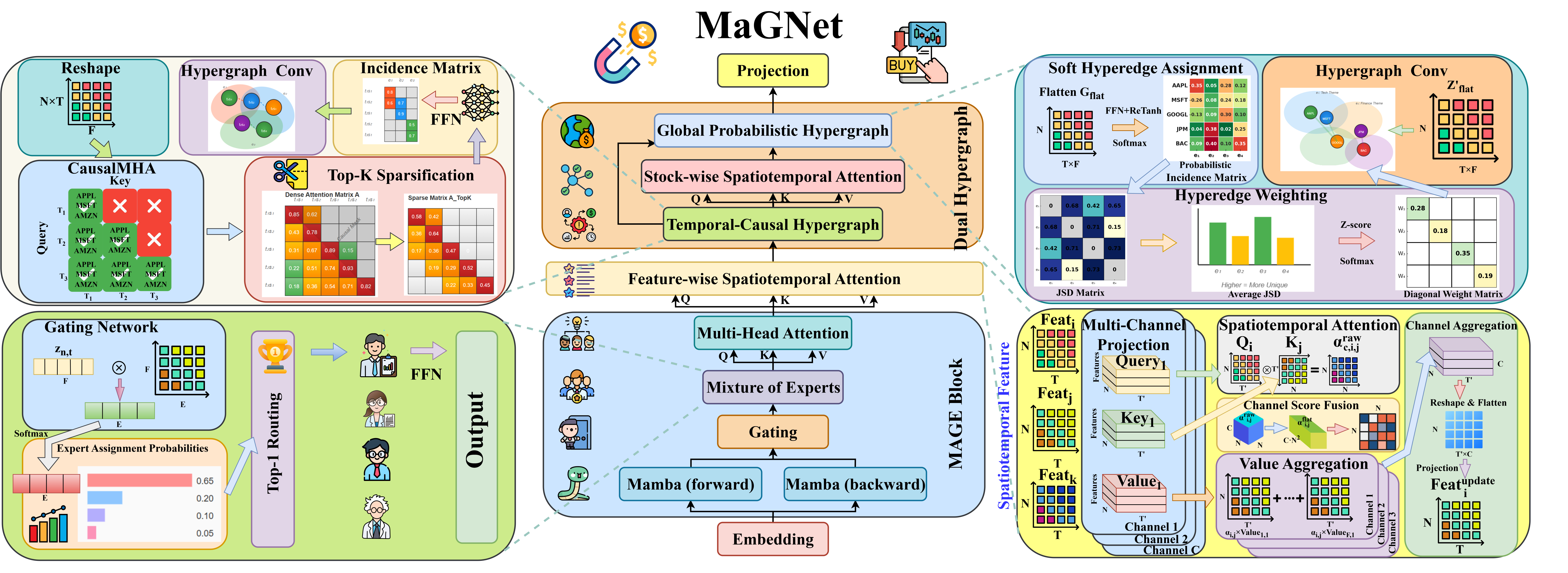}
\caption{Overview of MaGNet architecture. It consists of: (1) MAGE Block, combining bidirectional Mamba, adaptive gating, MoE, and multi-head attention for temporal modeling; (2) Feature-wise 2D Spatiotemporal Attention to capture cross-feature dependencies while preserving spatiotemporal structure; (3) Dual Hypergraph Module, with TCH modeling fine-grained temporal-causal relations and GPH capturing global market patterns via soft assignments and JSD-based weighting.}
\label{Overview of the MaGNet}
\end{figure*}

\subsection{Feature Embedding}
To compress information and extract salient features, each stock's daily feature vector $\mathbf{x}_{n,t} \in \mathbb{R}^{F}$ is first projected into a shared latent space through a feed-forward embedding network:
\begin{equation}
\mathbf{z}_{n,t} = \mathrm{Embedding}(\mathbf{x}_{n,t}) \in \mathbb{R}^{D}.
\end{equation}

\subsection{MAGE Block}
Figure \ref{Overview of the MaGNet} illustrates the overall architecture of MaGNet, with the \textbf{MAGE} (\textbf{M}amba-\textbf{A}ttention-\textbf{G}ating-\textbf{E}xperts) block serving as its core temporal modeling component. The MAGE block processes the embedded representations to capture heterogeneous temporal interactions through four synergistic modules: bidirectional Mamba first captures forward and backward temporal patterns, followed by a gating mechanism that adaptively fuses these contextual representations; then passes through a sparse Mixture of Experts layer for specialized processing under different market regimes; and finally, multi-head self-attention incorporates global dependencies across the entire sequence. This design balances computational efficiency with modeling capacity for complex market dynamics.

\subsubsection{Bidirectional Mamba}
Mamba \cite{gu2023mamba} enhances SSMs with selective scan mechanism and data-dependent parameters, achieving near-linear complexity while modeling long-range dependencies without requiring positional encodings \cite{park2024can}. However, standard Mamba processes sequences unidirectionally, limiting contextual understanding. We employ bidirectional Mamba \cite{ahamed2024timemachine} to address this:
\begin{equation}
\mathbf{Z}_n^{\mathrm{fwd}} = \mathrm{Mamba}(\mathbf{Z}_n),
\end{equation}
\begin{equation}
\mathbf{Z}_n^{\mathrm{bwd}} = \mathrm{Reverse}(\mathrm{Mamba}(\mathrm{Reverse}(\mathbf{Z}_n))).
\end{equation}
where $\mathbf{Z}_n = [\mathbf{z}_{n,1}, \mathbf{z}_{n,2}, \ldots, \mathbf{z}_{n,T}] \in \mathbb{R}^{T \times D}$ and $\mathrm{Reverse}(\cdot)$ denotes time-axis reversal. This bidirectional processing captures both forward and backward temporal dependencies, providing more comprehensive sequence representations essential for complex financial patterns.

\subsubsection{Gating Mechanism}

To fuse the forward and backward representations $\mathbf{Z}_n^{\mathrm{fwd}}$ and $\mathbf{Z}_n^{\mathrm{bwd}}$ into a unified encoding while avoiding naive averaging or concatenation, we employ a Gating Mechanism \cite{gru} that adaptively integrates these two directional outputs. At time step $t$:
\begin{align}
\mathbf{z}_{t}^{\mathrm{G}} &= \mathrm{Gate}(\mathbf{z}_{t}^{\mathrm{fwd}}, \mathbf{z}_{t}^{\mathrm{bwd}}) \\
 &= \sigma(W_{f}\mathbf{z}_{t}^{\mathrm{fwd}} + b_{f} + W_{b}\mathbf{z}_{t}^{\mathrm{bwd}} + b_{b}),
\end{align}
where $W_f, W_b \in \mathbb{R}^{D \times D}, b_f, b_b \in \mathbb{R}^D$ are learnable parameters.
This Gating Mechanism adaptively controls how much information to retain from each direction, enabling context-aware fusion for better temporal representation.

\subsubsection{Mixture of Experts}
Characterized by intrinsic volatility and structural heterogeneity, stock market environments pose significant challenges for distributionally robust modeling. Training data may originate from bullish markets, whereas evaluation could occur under bearish conditions or amid abrupt distributional shifts—such as major policy changes or global events. To address this, we incorporate a Switched Mixture-of-Experts (MoEs) \cite{shazeer2017outrageously} layer with $E$ experts that account for a wide range of possible market scenarios, enabling sparse conditional computation and enhancing model resilience without incurring scaling cost. First, the gating network computes expert assignment probabilities:
\begin{equation}
\mathbf{P} = \mathrm{Softmax}(\mathbf{Z}^{\mathrm{G}} \mathbf{W}_{\mathrm{g}}) \in \mathbb{R}^{N \times T \times E},
\end{equation}
where $\mathbf{W}_{\mathrm{g}} \in \mathbb{R}^{D \times E}$. 
We employ Top-1 routing using capacity-based normalization with scaling factor $C$ to enforce sparse expert selection and balanced utilization:
\begin{equation}
\hat{p}_{n,t,e} = 
\begin{cases}
p_{n,t,e}, & \text{if } e = \arg\max_j p_{n,t,j} \\
0, & \text{otherwise}
\end{cases},
\end{equation}
\begin{equation}
\tilde{p}_{n,t,e} = C \cdot \frac{\hat{p}_{n,t,e}}{\sum_{n',t'} \hat{p}_{n',t',e}}.
\end{equation}
Each expert $\mathcal{E}_e$ is a position-wise feed-forward network:
\begin{equation}
\mathbf{z}_{n,t}^{\mathrm{MoE}} = \mathcal{E}_{e^*}(\mathbf{z}_{n,t}^{\mathrm{G}}), \quad e^* = \arg\max_j p_{n,t,j},
\end{equation}
\begin{equation}
\mathcal{E}_e(\mathbf{z}) = \mathbf{W}_e^{(2)} \, \mathrm{GELU}(\mathbf{W}_e^{(1)} \mathbf{z} + \mathbf{b}_e^{(1)}) + \mathbf{b}_e^{(2)}.
\end{equation}
This design enables dynamic capacity allocation across time steps while maintaining load balancing through the capacity normalization mechanism.

\subsubsection{Multi-Head Self-Attention}

Finally, to complement bidirectional representations with global dependencies across the entire sequence, we incorporate multi-head attention \cite{attention}. Given input $\mathbf{Z}^{\mathrm{MoE}} \in \mathbb{R}^{N \times T \times D}$, we compute $h$ attention heads in parallel across multiple representation subspaces with size $d_h = D/h$ and combine them:
\begin{equation}
\mathbf{Q}_i = \mathbf{Z}^{\mathrm{MoE}} \mathbf{W}_i^Q, \enspace
\mathbf{K}_i = \mathbf{Z}^{\mathrm{MoE}} \mathbf{W}_i^K, \enspace
\mathbf{V}_i = \mathbf{Z}^{\mathrm{MoE}} \mathbf{W}_i^V,
\end{equation}
\begin{equation}
\mathrm{head}_i = \mathrm{Softmax}\left(\frac{\mathbf{Q}_i \mathbf{K}_i^\top}{\sqrt{d_k}}\right) \mathbf{V}_i,
\end{equation}
\begin{equation}
\mathbf{Z}^{\mathrm{MAGE}} = \mathrm{Concat}(\mathrm{head}_1, \ldots, \mathrm{head}_h) \mathbf{W}^O,
\end{equation}
where $\mathbf{W}_i^Q, \mathbf{W}_i^K, \mathbf{W}_i^V \in \mathbb{R}^{D \times d_h}, \forall i \in \{1, \ldots, h\}$ and $\mathbf{W}^O \in \mathbb{R}^{D \times D}$. This module enables the model to jointly attend to information from different representation subspaces, effectively capturing global dependencies.

Together, these four components form a flexible and powerful architecture capable of learning diverse temporal patterns in financial time series.

\subsection{Feature-wise 2D Spatiotemporal Attention}
Traditional time series methods focus on single dimension (typically temporal), requiring multivariate data to be flattened or processed separately, disrupting spatiotemporal structure and may obscure meaningful relationships across time and assets. While iTransformer \cite{liu2023itransformer} models cross-feature dependencies, it overlooks the crucial spatiotemporal relationships. We propose Feature-wise 2D Spatiotemporal Attention that preserves this structure by representing each feature as an $N \times T$ matrix (stocks × time), enabling direct feature-to-feature interactions while maintaining spatiotemporal integrity, yielding rich representations for downstream dual hypergraph learning.

The module first transposes $\mathbf{Z}^{\mathrm{MAGE}} \in \mathbb{R}^{N \times T \times D}$ into $\mathbf{Z}' \in \mathbb{R}^{D \times N \times T}$, where each feature $d$ is represented by matrix $\mathbf{Z}'_d \in \mathbb{R}^{N \times T}$, capturing dynamics across all stocks and time steps. To handle the increased element dimensionality while maintaining computational efficiency, we employ $C$ parallel attention channels. For each feature $d$ and channel $c \in \{1, \ldots, C\}$, we compute query, key and value projections:
$
\mathbf{Q}_{d,c} = \mathbf{Z}'_d \mathbf{W}_c^Q + \mathbf{b}_c^Q, \enspace
\mathbf{K}_{d,c} = \mathbf{Z}'_d \mathbf{W}_c^K + \mathbf{b}_c^K, \enspace
\mathbf{V}_{d,c} = \mathbf{Z}'_d \mathbf{W}_c^V + \mathbf{b}_c^V,
$
where $\mathbf{W}_c^Q, \mathbf{W}_c^K, \mathbf{W}_c^V \in \mathbb{R}^{T \times T'}$ and $\mathbf{b}_c^Q, \mathbf{b}_c^K, \mathbf{b}_c^V \in \mathbb{R}^{1 \times T'}$ are learnable parameters. To capture feature-to-feature relationships, we compute cross-stock multi-channel attention scores between each feature pair $i, j \in \{1, \ldots, D\}$, and then fuse them through a feed-forward network:
\begin{equation}
\boldsymbol{\alpha}_{i,j}^{\mathrm{raw}} = \frac{1}{\sqrt{T'}} \left[\mathbf{Q}_{i,1} \mathbf{K}_{j,1}^\top, \ldots, \mathbf{Q}_{i,C} \mathbf{K}_{j,C}^\top\right] \in \mathbb{R}^{C \times N \times N},
\end{equation}
\begin{equation}
\boldsymbol{\alpha}_{i,j}^{\mathrm{flat}} = \mathrm{Flatten}(\boldsymbol{\alpha}_{i,j}^{\mathrm{raw}}) \in \mathbb{R}^{C \cdot N^2}, \alpha'_{i,j} = \mathrm{FFN}(\boldsymbol{\alpha}_{i,j}^{\mathrm{flat}}) \in \mathbb{R}.
\end{equation}
The aggregated representations are computed via weighted summation:
\begin{equation}
\mathbf{B}'_{i,c} = \sum_{j=1}^{D} a_{i,j} \mathbf{V}_{j,c} \in \mathbb{R}^{N \times T'}, \enspace a_{i,j} = \mathrm{softmax}(\alpha'_{i,j}).
\end{equation}
All channels' outputs $\{\mathbf{B}'_{i,c}\}_{c=1}^C$ are then stacked into $\mathbf{B}_i' \in \mathbb{R}^{C \times N \times T'}$, reshaped to $\mathbf{B}_i \in \mathbb{R}^{N \times (T' \cdot C)}$, and projected to dimension $T$ via feed-forward network to obtain $\mathbf{z}_i^{\mathrm{F2D}} \in \mathbb{R}^{N \times T}$. Stacking all features yields $\mathbf{Z}^{\mathrm{F2D}} \in \mathbb{R}^{N \times T \times D}$ with enriched cross-feature dependencies.

\subsection{Dual-Hypergraph Learning}
Traditional graph methods capture only pairwise relationships, missing the group dynamics where multiple stocks move synchronously. Hypergraphs address this by connecting multiple nodes via hyperedges. We propose a dual hypergraph framework leveraging the spatiotemporal representations from our 2D attention mechanisms: the Temporal-Causal Hypergraph (TCH) learns dynamic causal relationships respecting temporal ordering, while the Global Probabilistic Hypergraph (GPH) captures instantaneous market-wide patterns through soft hyperedge assignments. This design models both localized temporal influences and global market structures for multi-scale market interdependencies.

\subsubsection{Temporal-Causal Hypergraph (TCH)}
The Temporal-Causal Hypergraph (TCH) discovers high-order causal dependencies among stocks at the fine-grained stock-time level—where each stock at a specific timestamp can influence multiple others at current or subsequent timestamps. Unlike static graph approaches, TCH adaptively learns dynamic hypergraph topology that respects temporal ordering, tracking how stock movements propagate through the market over time. We restructure the input $\mathbf{Z}^{\mathrm{F2D}}$ by flattening:
\begin{equation}
\mathbf{Z}_{\mathrm{flat}} = \mathrm{Flatten}((\mathbf{Z}^{\mathrm{F2D}})^\top) \in \mathbb{R}^{(T \cdot N) \times D},
\end{equation}
where each row represents a time-stock pair. We then apply Causal Multi-Head Attention (CausalMHA) to capture causal relationships while preventing future information leakage:
\begin{equation}
\mathbf{A} = \mathrm{CausalMHA}(\mathbf{Z}_{\mathrm{flat}}) \in \mathbb{R}^{(T \cdot N) \times (T \cdot N)}.
\end{equation}
The causality constraint is enforced through a upper triangular block mask such that each node can only attend to others from current or earlier time steps. This explicit design captures temporal and cross-sectional dependencies with causal validity, modeling how past conditions and peer movements influence future stock behavior.

The dense attention matrix $\mathbf{A}$ may contain weak or noisy connections that can impair performance. To extract the most salient relationships, we apply Top-$K$ sparsification:
\begin{equation}
\mathbf{A}_{\mathrm{TopK}}[i,j] = 
\begin{cases}
\mathrm{Softmax}(\mathbf{A}[i,:])_j,& \text{if } j \in \mathrm{Top}\text{-}K(i) \\
0, & \text{otherwise}
\end{cases}.
\end{equation}
This sparsification highlights the strongest causal paths, reduces computational complexity, and mitigates the influence of noisy or insignificant connections.

To transform pairwise attention scores into the hypergraph structure, we introduce a FFN with a novel Rectified Tanh (ReTanh) activation:
\begin{equation}
    \mathbf{Z}_1 = \mathrm{ReTanh}(\mathbf{W}_1 \mathbf{A}_{\mathrm{TopK}}), 
\mathbf{H}_{\mathrm{TCH}} = \mathrm{ReTanh}(\mathbf{W}_2 \mathbf{Z}_1),
\end{equation}
where $\mathbf{H}_{\mathrm{TCH}} \in \mathbb{R}^{(T \cdot N) \times M_1}$ is incidence matrix, $M_1$ is the number of hyperedges. The ReTanh activation is defined as:
\begin{equation}
\mathrm{ReTanh}(x) = 
\begin{cases}
0 & \text{if } x \leq 0 \\
\tanh(x) & \text{if } x > 0
\end{cases}.
\end{equation}
ReTanh enhances hypergraph learning by combining sparsity through rectified filtering of weak linkages with stability from tanh-bounded hyperedge assignments, reducing outlier impact like abnormal financial events, and ensuring robust, balanced correlations in noisy market environments.

Next, we perform hypergraph convolution to enable high-order information propagation:
\begin{equation}
\mathbf{Z}'_{\mathrm{flat}} = \mathrm{ELU}\left(\mathbf{H}_{\mathrm{TCH}} \mathbf{H}_{\mathrm{TCH}}^\top \mathbf{Z}_{\mathrm{flat}} \mathbf{P}_1\right),
\end{equation}
where $\mathbf{P}_1 \in \mathbb{R}^{D \times D}$ is a learnable projection matrix. This operation enables each node to aggregate information from all nodes within its hyperedges, capturing complex group dynamics that pairwise methods miss. Finally, we reshape the transformed features back to the original format $\mathbf{Z}^{\mathrm{TCH}} = \mathrm{Reshape}(\mathbf{Z}'_{\mathrm{flat}}) \in \mathbb{R}^{N \times T \times D}$.

TCH achieves adaptive learning of temporal-causal relationships that respect market dynamics. The combination of causal attention for temporal validity and hypergraph convolution for high-order modeling creates a powerful framework for capturing the complex interdependencies that drive stock market movements.

\subsubsection{Global Probabilistic Hypergraph (GPH)}
While TCH captures temporal-causal relationships, financial markets also exhibit instantaneous global patterns. The Global Probabilistic Hypergraph (GPH) discovers these market-wide group interactions through probabilistic hyperedges, allowing stocks to participate in multiple market themes simultaneously with varying membership degrees.

To model direct inter-stock relationships, we first introduce Stock-wise 2D Spatiotemporal Attention, applying similar mechanism as its feature-wise counterpart but along the stock dimension, where each stock $n$ is represented by its temporal features $\mathbf{Z}_n^{\mathrm{TCH}} \in \mathbb{R}^{T \times D}$, producing output $\mathbf{Z}^{\mathrm{N2D}} \in \mathbb{R}^{N \times T \times D}$ that encodes cross-stock dependencies.

To learn soft hyperedge assignments, we flatten $\mathbf{Z}^{\mathrm{N2D}}$ into $\mathbf{G}_{\mathrm{flat}} \in \mathbb{R}^{N \times (T \cdot D)}$, apply a FFN with ReTanh activation, and normalize column-wise to obtain the probabilistic incidence matrix $\mathbf{H}_{\mathrm{GPH}} \in \mathbb{R}^{N \times M_2}$:
\begin{equation}
\mathbf{H}_{\mathrm{GPH}} = \mathrm{softmax}\left( \mathrm{ReTanh}\left( \mathrm{FFN}(\mathbf{G}_{\mathrm{flat}}) \right)\right),
\end{equation}
where $M_2$ is the number of hyperedges. Each column $\mathbf{e}_j$ in $\mathbf{H}_{\mathrm{GPH}}$ defines a soft hyperedge as a probability distribution with membership probabilities summing to 1.

To address redundant relationships between hyperedges, we weight each hyperedge by its distinctiveness using Jensen-Shannon Divergence (JSD). For each hyperedge, we compute its average divergence:
\begin{equation}
\mu_j = \frac{1}{M_2} \sum_{i=1}^{M_2} \mathrm{JSD}(\mathbf{e}_i \parallel \mathbf{e}_j),
\end{equation}
JSD's symmetry and boundedness ($[0, \log 2]$) ensure fair evaluation regardless of comparison order and stable optimization. The importance weight is:
\begin{equation}
w_j = \mathrm{softmax}(\mathrm{Z\text{-}score}(\mu_j)),
\end{equation}
where Z-score prevents domination by only a small number of hyperedges. This JSD weighting scheme assigns higher weights to unique hyperedges that capture distinct and informative market structures.

Finally, the global hypergraph convolution integrates the weighted hyperedges to propagate information across all stocks:
\begin{equation}
\mathbf{Z}^{\mathrm{GPH}} = \mathrm{ELU}(\mathbf{H}_{\mathrm{GPH}} \mathbf{W} \mathbf{H}_{\mathrm{GPH}}^\top \mathbf{Z}'_{\mathrm{flat}} \mathbf{P}_2) \in \mathbb{R}^{N \times (T \cdot D)},
\end{equation}
where $\mathbf{W} = \mathrm{diag}(w_1, \ldots, w_{M_2})$ is diagonal weight matrix and $\mathbf{P}_2 \in \mathbb{R}^{(T \cdot D) \times (T \cdot D)}$ is a learnable projection matrix.

GPH propagates information globally via weighted group memberships and importance scores. Combined with TCH's temporal-causal patterns, this dual hypergraph framework captures both localized temporal influences and global market structures, providing the multi-scale representational capacity essential for accurate stock prediction.

\section{Experiments}
\subsection{Datasets}
We evaluated our method on six major stock indices (DJIA, HSI, NASDAQ 100, S\&P 100, CSI 300, and Nikkei 225) using data from January 1, 2020 through December 31, 2024. Data was split 7:1:2 for training, validation (hyperparameter selection) and testing (evaluation). See Table \ref{tab:datasets} for dataset statistics.

\begin{table}[tbh!]
\begin{center}
\caption{Statistics of Datasets}
\label{tab:datasets}
\begin{tabular}{ccccc}
\toprule
\textbf{Dataset} & \textbf{\# Stocks} & \textbf{\# Training} & \textbf{\# Val} & \textbf{\# Test} \\ \midrule
\textbf{DJIA}       & 30  & 879 & 125 & 253 \\ 
\textbf{HSI}        & 71  & 860 & 122 & 247 \\ 
\textbf{NASDAQ 100} & 92  & 879 & 125 & 253 \\ 
\textbf{S\&P 100}    & 99  & 879 & 125 & 253 \\ 
\textbf{CSI 300}    & 215 & 848 & 121 & 243 \\ 
\textbf{Nikkei 225} & 222 & 855 & 122 & 245 \\ \bottomrule
\end{tabular}
\end{center}

\end{table}

\begin{table*}[htb]
\begin{center}
\caption{Prediction performance comparison on DJIA, HSI, and NASDAQ 100. The best results are in bold and the second-best results are underlined.}
\label{tab:Prediction results 1}
\setlength{\tabcolsep}{1mm}
\begin{tabular}{cccccccccccccccc}
\toprule
\multicolumn{1}{c}{\multirow{2}{*}{Model}} & \multicolumn{5}{c}{DJIA}                                                            & \multicolumn{5}{c}{HSI}                                                            & \multicolumn{5}{c}{NASDAQ 100}                                                     \\ \cmidrule(lr){2-6} \cmidrule(lr){7-11} \cmidrule(lr){12-16}
\multicolumn{1}{c}{}                       & ACC            & PRE            & REC             & F1             & AUC            & ACC            & PRE            & REC            & F1             & AUC            & ACC            & PRE            & REC            & F1             & AUC            \\ \midrule
GRU                                        & 51.59          & 53.13          & 75.91           & 62.51          & 49.98          & 51.89          & 49.12          & 1.73           & 3.34           & 50.78          & 52.07          & 52.37          & 88.86          & 65.90          & 51.41          \\
LSTM                                       & 53.03          & 53.15          & 98.17           & 68.97          & 50.57          & 51.96          & 50.35          & 7.11           & 12.46          & 50.75          & 51.86          & 52.09          & 95.07          & 67.30          & 50.52          \\
DLinear                                    & 52.39          & 52.98          & 92.81           & 67.46          & 49.65          & 52.20          & 51.46          & 10.25          & 17.09          & 50.78          & 52.18          & 52.25          & 95.58          & 67.57          & 51.32          \\
TimesNet                                   & 50.34          & 52.53          & 68.43           & 59.44          & 49.04          & 52.40          & 50.92          & 27.99          & 36.13          & 52.07          & 51.77          & 52.18          & 89.32          & 65.88          & 50.37          \\
PatchTST                                   & 49.66          & 52.62          & 53.15           & 52.89          & 49.04          & 50.25          & 47.44          & 32.13          & 38.31          & 49.40          & 51.72          & 52.23          & 86.44          & 65.11          & 49.99          \\
iTransformer                                    & {\ul 53.13}    & {\ul 53.39}    & 93.30           & 67.91          & 50.11          & 52.50          & 51.22          & 25.49          & 34.03          & 52.32          & 51.09          & 51.76          & 90.76          & 65.92          & 48.57          \\
TimeM.                                     & 52.39          & 53.02          & 91.55           & 67.15          & 49.36          & 51.89          & 49.87          & 11.42          & 18.58          & 51.13          & {\ul 52.64}    & 52.53          & 94.88          & 67.62          & 51.93          \\
TimeXer                                     & 52.92          & 53.07          & 98.71           & 69.03          & 49.06          & 52.03          & 50.90          & 6.65           & 11.76          & 50.84          & 52.50          & 52.43          & {\ul 95.69}    & {\ul 67.74}    & {\ul 52.12}    \\
GCN                                        & 52.30          & 53.05          & 89.50           & 66.61          & 49.65          & 51.83          & 49.81          & 24.92          & 33.22          & 51.52          & 52.00          & {\ul 52.91}    & 71.86          & 60.95          & 50.86          \\
GraphSAGE                                     & 52.88          & 53.03          & {\ul 99.25}     & {\ul 69.13}    & 49.59          & {\ul 52.76}    & 51.40          & 32.13          & 39.55          & 52.87          & 51.47          & 52.19          & 82.28          & 63.86          & 51.93          \\
GAT                                        & 52.90          & 53.11          & 97.39           & 68.74          & 50.03          & 52.32          & 51.39          & 15.56          & 23.89          & {\ul 52.93}    & 52.37          & 52.41          & 93.86          & 67.26          & 50.05          \\
SFM                                        & 49.89          & \textbf{53.44} & 55.02           & 54.22          & 48.96          & 51.11          & 48.51          & {\ul 33.84}    & {\ul 39.87}    & 50.68          & 51.05          & 52.82          & 61.45          & 56.81          & 50.81          \\
Adv-ALSTM                                     & 52.69          & 53.07          & 95.06           & 68.11          & 50.72          & 51.07          & 48.06          & 21.94          & 30.08          & 49.92          & 51.90          & 52.14          & 94.61          & 67.21          & 49.83          \\
DTML                                       & 52.30          & 53.02          & 90.37           & 66.83          & {\ul 50.85}    & 51.73          & 49.69          & 31.91          & 38.87          & 50.82          & 52.01          & 52.52          & 82.49          & 64.18          & 51.05          \\
ESTIMATE                                     & 52.93          & 53.15          & 96.77           & 68.61          & 50.82          & 52.08          & 50.95          & 8.91           & 15.17          & 50.64          & 51.94          & 52.15          & 94.41          & 67.19          & 50.57          \\
StockMixer                                    & 52.49          & 53.10          & 90.90           & 67.04          & 49.16          & 51.92          & 50.00          & 15.59          & 23.76          & 50.84          & 51.70          & 52.02          & 94.68          & 67.14          & 49.23          \\
MASTER                                      & 52.15          & 53.18          & 83.53           & 64.98          & 49.78          & 52.58          & {\ul 51.88}    & 19.14          & 27.97          & 52.49          & 51.81          & 52.38          & 83.04          & 64.24          & 51.92          \\
MaGNet                                     & \textbf{53.16} & 53.16          & \textbf{100.00} & \textbf{69.42} & \textbf{51.10} & \textbf{54.19} & \textbf{52.85} & \textbf{43.90} & \textbf{47.96} & \textbf{54.12} & \textbf{53.72} & \textbf{53.09} & \textbf{96.14} & \textbf{68.41} & \textbf{52.24}\\ \bottomrule
\end{tabular}
\end{center}
\end{table*}

\subsection{Features}
We obtained historical stock data from Yahoo Finance \footnote{\url{https://ranaroussi.github.io/yfinance/}}, collecting five attributes: close, high, low, open, volume. To enhance features, we used Qlib \cite{yang2020qlib} to compute Alpha158 and Alpha360 technical indicators. After filtering out missing values, we combined these features with the five attributes to create an enriched dataset. We applied Z-Score normalization independently to each data split to prevent information leakage while maintaining stable training process.

\subsection{Baselines}
We evaluate our method against 17 baselines spanning three categories:
\begin{itemize}
    \item \textbf{Stock Prediction Models} (6): SFM \cite{zhang2017stock}, Adv-ALSTM \cite{feng2018enhancing}, DTML \cite{attention1}, ESTIMATE \cite{huynh2023efficient}, StockMixer \cite{fan2024stockmixer}, MASTER \cite{li2024master};
    \item \textbf{Time Series Models} (8): GRU \cite{gru}, LSTM \cite{gers2000learning}, DLinear \cite{zeng2023transformers}, TimesNet \cite{wu2022timesnet}, PatchTST \cite{nie2022time}, iTransformer \cite{liu2023itransformer}, TimeMixer \cite{wang2024timemixer}, TimeXer \cite{wang2024timexer};
    \item \textbf{Graph Models} (3): GCN \cite{kipf2016semi}, GraphSAGE \cite{hamilton2017inductive}, GAT \cite{velivckovic2017graph}.
\end{itemize}
Baselines' descriptions are provided in Appendix \ref{Baseline Descriptions}.

\subsection{Evaluation Metrics}
We evaluate model's performance using classification metrics and portfolio backtesting. For predictive abilities, we use Accuracy (ACC), Precision (PRE), Recall (REC), F1 score, and AUC. To assess model's profitability and risk in simulated investment scenarios, we employ Annual Return (AR), Sharpe Ratio (SR, applying a 2\% risk-free rate), Calmar Ratio (CR), and Maximum Drawdown (MDD). Detailed definitions and formulas of all metrics are provided in Appendix \ref{Metric Definitions}. 

\subsection{Hyperparameter Settings}
We passed the final representations $\mathbf{Z}^{\text{GPH}}$ through a feed-forward network followed by a softmax layer to predict the probability distribution of next-day stock movement directions. The model was implemented in PyTorch and optimized with cross-entropy loss. Hyperparameters were selected via grid search to maximize validation accuracy. Detailed configurations and dataset-specific settings are provided in Appendix \ref{sec:hypersetting}.

\begin{table*}[htb]
\centering
\caption{Backtesting performance on DJIA, HSI, and NASDAQ 100. The best results are in bold and the second-best results are underlined.}
\label{tab:Backtest results 1}
\begin{tabular}{ccccccccccccc}
\toprule
\multicolumn{1}{c}{\multirow{2}{*}{Model}} & \multicolumn{4}{c}{DJIA}                                       & \multicolumn{4}{c}{HSI}                                        & \multicolumn{4}{c}{NASDAQ 100}                                 \\  \cmidrule(lr){2-5} \cmidrule(lr){6-9} \cmidrule(lr){10-13}
\multicolumn{1}{c}{}                       & AR             & SR            & CR            & MDD           & AR             & SR            & CR            & MDD           & AR             & SR            & CR            & MDD           \\ \midrule
GRU                                        & 18.84          & 0.62          & 1.21          & 15.53         & 8.38           & 0.54          & 0.71          & 11.85         & 15.98          & 0.83          & {\ul 1.79}    & 8.92          \\
LSTM                                       & 17.00          & 1.41          & 3.30          & 5.15          & 5.32           & 0.25          & 0.44          & 11.97         & 11.68          & 0.65          & 1.16          & 10.10         \\
DLinear                                    & 8.81           & 0.64          & 1.45          & 6.07          & 7.49           & 0.48          & 1.02          & 7.38          & 10.26          & 0.56          & 1.04          & 9.83          \\
TimesNet                                   & 1.99           & -0.01         & \textbf{7.95} & \textbf{0.25} & 6.43           & 0.50          & 1.18          & 5.45          & 6.36           & 0.29          & 0.59          & 10.83         \\
PatchTST                                   & 1.28           & -0.22         & 0.63          & {\ul 2.04}    & 5.31           & \textbf{0.70} & \textbf{2.69} & {\ul 1.97}    & -3.08          & -0.35         & -0.24         & 12.94         \\
iTransformer                                    & 9.82           & 0.74          & 1.83          & 5.37          & 5.45           & 0.40          & 1.00          & 5.45          & 2.16           & 0.01          & 0.20          & 10.70         \\
TimeMixer                                     & 9.80           & 0.74          & 1.68          & 5.82          & 7.02           & 0.51          & 0.94          & 7.49          & 12.30          & 0.70          & 1.20          & 10.21         \\
TimeXer                                     & 17.68          & 1.48          & 3.43          & 5.16          & 7.53           & 0.21          & 0.40          & 18.65         & 11.90          & 0.66          & 1.18          & 10.10         \\
GCN                                        & 11.99          & 0.88          & 1.71          & 7.00          & 8.25           & 0.43          & 0.80          & 10.37         & 7.47           & 0.43          & 0.87          & 8.60          \\
GraphSAGE                                     & 18.29          & {\ul 1.55}    & 3.35          & 5.46          & 9.91           & 0.56          & 0.95          & 10.41         & 4.72           & 0.23          & 0.55          & 8.53          \\
GAT                                        & 6.38           & 0.40          & 1.09          & 5.83          & 4.28           & 0.16          & 0.43          & 9.85          & 10.65          & 0.59          & 1.14          & 9.37          \\
SFM                                        & 6.88           & 0.70          & 1.73          & 3.98          & 2.09           & 0.02          & {\ul 2.02}    & \textbf{1.04} & 2.85           & 0.14          & 0.67          & \textbf{4.23} \\
Adv-ALSTM                                     & 15.01          & 1.23          & 2.92          & 5.14          & 3.15           & 0.10          & 0.30          & 10.38         & 11.03          & 0.62          & 1.29          & 8.55          \\
DTML                                       & 13.82          & 1.11          & 2.71          & 5.09          & 8.11           & 0.35          & 0.68          & 11.94         & 7.11           & 0.34          & 0.70          & 10.19         \\
ESTIMATE                                     & \textbf{24.40} & 1.42          & 3.38          & 7.21          & \textbf{16.26} & {\ul 0.66}    & 1.01          & 16.12         & {\ul 16.77}    & {\ul 1.03}    & 1.78          & 9.44          \\
StockMixer                                    & 8.46           & 0.61          & 1.47          & 5.76          & 4.74           & 0.30          & 1.00          & 4.75          & 10.02          & 0.54          & 1.00          & 9.97          \\
MASTER                                      & 3.40           & 0.13          & 0.52          & 6.50          & 5.19           & 0.20          & 0.46          & 11.33         & 7.03           & 0.44          & 0.83          & 8.43          \\
MaGNet                                     & {\ul 19.92}    & \textbf{1.70} & {\ul 3.93}    & 5.07          & {\ul 12.25}    & {\ul 0.66}    & 1.33          & 9.20          & \textbf{17.09} & \textbf{1.05} & \textbf{2.09} & {\ul 8.18}   \\ \bottomrule
\end{tabular}
\end{table*}

\begin{table*}[htb]
\begin{center}
\caption{Ablation results on NASDAQ 100. The best results are in bold and the second-best results are underlined.}
\label{tab:AblationNA}
\begin{tabular}{cllllllllll}
\toprule
\multirow{2}{*}{Dataset}    & \multicolumn{1}{c}{\multirow{2}{*}{Component}} & \multicolumn{5}{c}{Prediction}                                                      & \multicolumn{4}{c}{Backtesting}                                \\ \cmidrule(lr){3-7} \cmidrule(lr){8-11}
                            & \multicolumn{1}{c}{}                           & ACC            & PRE            & REC             & F1             & AUC            & AR             & SR            & CR            & MDD           \\ \midrule

\multirow{5}{*}{NASDAQ 100} & w/o MAGE                                       & {\ul 52.97}    & {\ul 52.90}    & 88.98           & 66.36          & 50.29          & 9.68           & 0.56          & 1.09          & {\ul 8.85}    \\
                            & w/o F. 2D Attn                                 & 52.17          & 52.26          & 95.11           & 67.46          & 48.46          & {\ul 13.41}    & {\ul 0.80}    & {\ul 1.35}    & 9.96          \\
                            & w/o TCH                                        & 52.73          & 52.56          & {\ul 95.36}     & {\ul 67.77}    & {\ul 50.97}    & 9.49           & 0.50          & 0.92          & 10.31         \\
                            & w/o GPH                                        & 52.65          & 52.61          & 92.16           & 66.99          & 47.98          & 8.44           & 0.48          & 0.78          & 10.76         \\
                            & MaGNet                                         & \textbf{53.72} & \textbf{53.09} & \textbf{96.14}  & \textbf{68.41} & \textbf{52.24} & \textbf{17.09} & \textbf{1.05} & \textbf{2.09} & \textbf{8.18} \\ \bottomrule
\end{tabular}
\end{center}
\end{table*}

\subsection{Prediction Results}
The prediction results on DJIA, HSI and NASDAQ 100 indices are shown in Table \ref{tab:Prediction results 1}, and results on S\&P 100, CSI300 and Nikkei 255 are presented in Appendix \ref{Additional Prediction Results}. MaGNet consistently outperforms all baselines across all indices in terms of accuracy, recall, F1-score, and AUC, demonstrating its superior predictive capability. Notably, it achieves the highest accuracy on HSI (54.19\%) and competitive performance on DJIA and NASDAQ 100. The model also delivers substantial improvements in recall, particularly for DJIA, indicating strong sensitivity to upward movements. Across these datasets, MaGNet maintains balanced precision–recall trade-offs, reflected in leading or near-leading F1-scores. Its consistently higher AUC values further suggest enhanced discrimination ability compared to both time-series-only and graph-based baselines.

\subsection{Backtesting Results}

To evaluate the practical profitability of the proposed model in realistic trading scenarios, we conduct daily backtests using a systematic portfolio-based trading strategy that simulates real-world trading mechanics. The detailed description of the trading strategy and the configurations can be found in Appendix \ref{sec:backtest}.

The backtesting results on DJIA, HSI, and NASDAQ 100 indices are summarized in Table \ref{tab:Backtest results 1}, and results on the other three indices are provided in Appendix \ref{Additional Backtesting Results}. Across these indices, MaGNet consistently converts predictive gains into higher risk-adjusted returns, achieving the highest or near-highest Sharpe Ratios and Annual Returns among all models. Drawdowns remain moderate compared with baselines, reflecting effective downside control. These findings demonstrate that integrating MAGE’s temporal modeling with the dual-hypergraph framework yields robust and profitable trading performance across various indices.

\begin{figure*}[htbp]
    \centering
    \includegraphics[width=\textwidth]{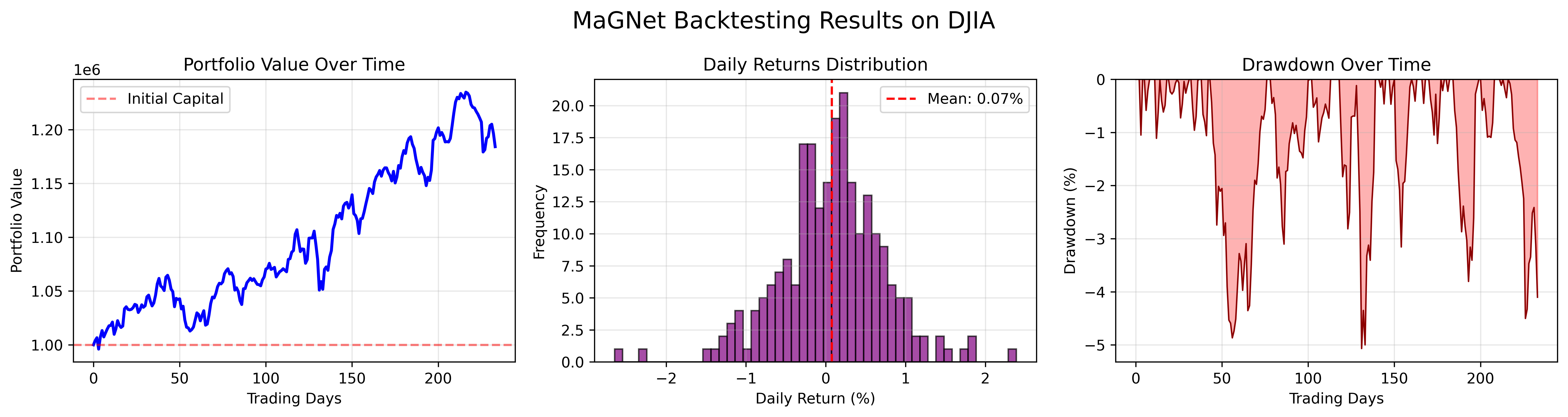}
    \includegraphics[width=\textwidth]{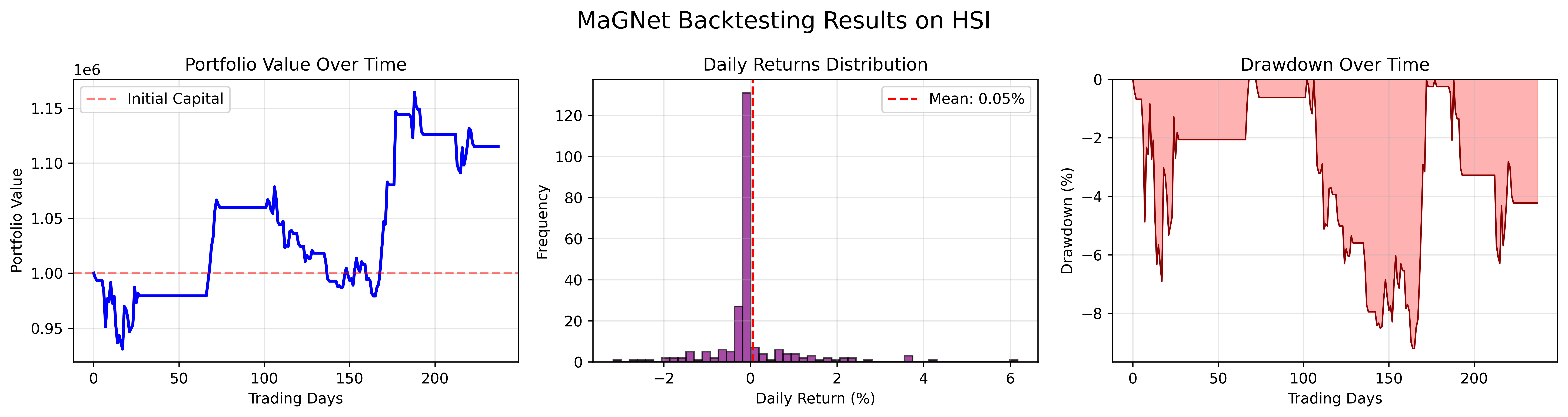}
    \includegraphics[width=\textwidth]{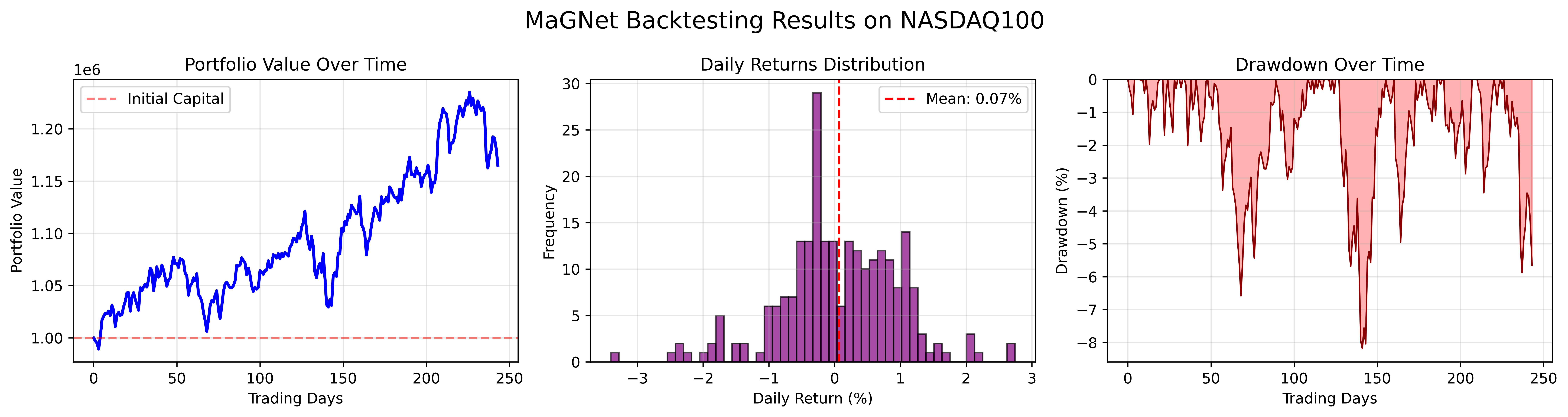}
    \caption{Backtesting performance of the MaGNet on DJIA, HSI and NASDAQ 100 indices
    }
    \label{fig:backtesting_results1}
\end{figure*}

\begin{figure*}[htbp]
    \centering
    \includegraphics[width=\textwidth]{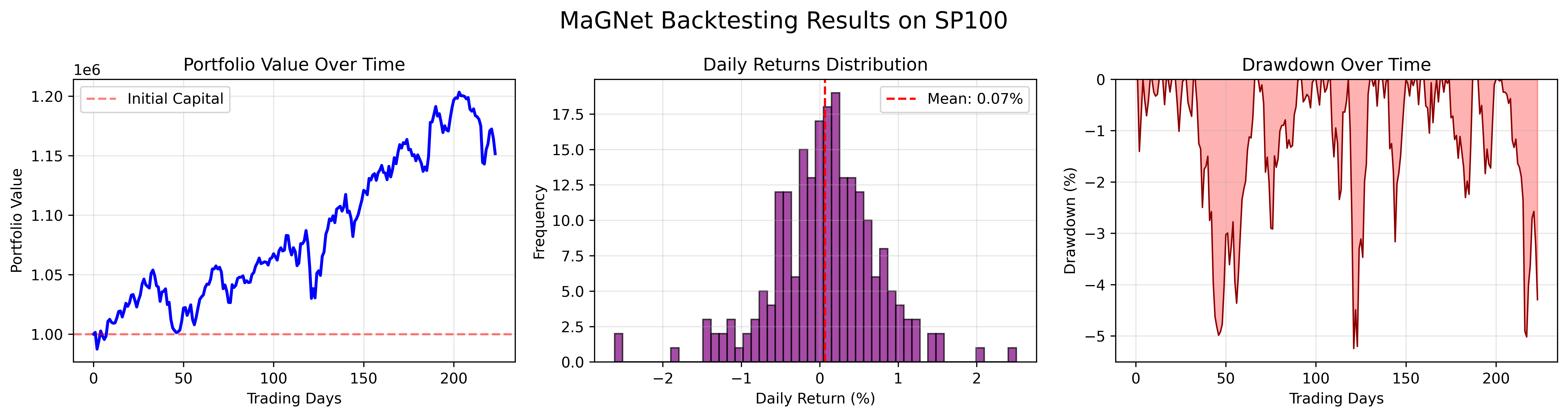}
    \includegraphics[width=\textwidth]{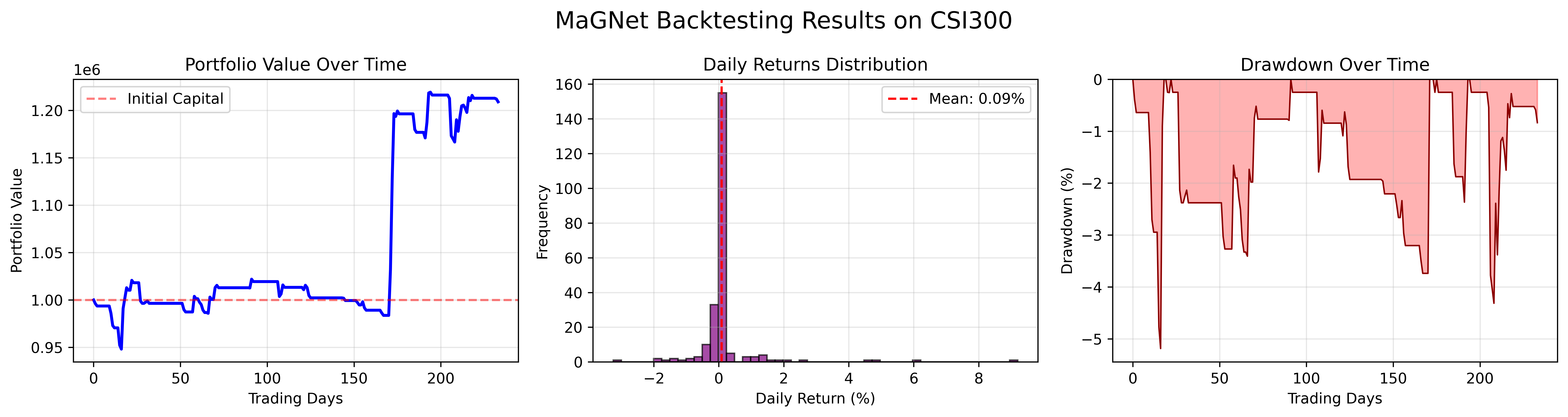}
    \includegraphics[width=\textwidth]{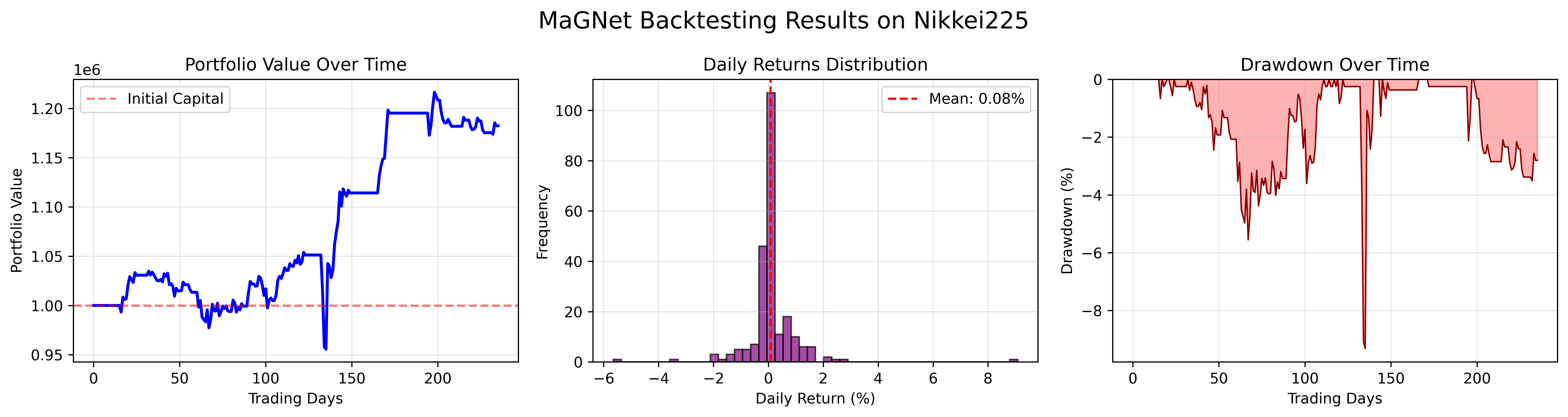}
    \caption{Backtesting performance of the MaGNet on S\&P 100, CSI300 and Nikkei225 indices
    }
    \label{fig:backtesting_results2}
\end{figure*}

Figure \ref{fig:backtesting_results1} and \ref{fig:backtesting_results2} present the backtesting performance visualization of MaGNet across all six stock indices, with each subplot containing three panels: portfolio value over time (left), daily returns distribution (center), and drawdown trajectory (right). The portfolio value trajectories demonstrate MaGNet's consistent ability to generate positive returns, with particularly strong performance on CSI300 (achieving approximately 22\% growth), DJIA (20\% growth), and Nikkei225 (18\% growth). The daily returns distributions exhibit near-normal characteristics centered slightly above zero (mean daily return of 0.07-0.09\%), indicating the model's capability to maintain positive expected returns while avoiding extreme tail events. Notably, the drawdown profiles reveal MaGNet's robust risk management, with maximum drawdowns contained below 5\% for CSI300 and DJIA, and remaining under 10\% across all indices. The relatively shallow and quickly-recovering drawdown patterns, particularly evident in the DJIA and CSI300 results, corroborate the superior Sharpe ratios reported in Table \ref{tab:Backtest results 1} and \ref{tab:Backtest results 2}, demonstrating that MaGNet not only achieves strong absolute returns but does so with controlled downside risk, a critical requirement for practical trading applications.

\subsection{Ablation Studies}
To evaluate the contribution of each component, we conducted ablation studies by removing key modules from MaGNet. The complete ablation results across all six indices are provided in Appendix \ref{Complete Results of Ablation Studies}. The results on NASDAQ100 are shown in Table \ref{tab:AblationNA}. The results indicate that each component contributes critically to MaGNet’s performance. Removing the MAGE block causes the most substantial degradation in trading performance—Annual Return drops from 17.09\% to 8.44\%, confirming MAGE’s essential role in capturing complex temporal dynamics for both prediction and profitability. The Feature-wise 2D Spatiotemporal Attention and the dual hypergraph components also play important roles, demonstrating that MaGNet’s superior performance arises from the integration of advanced temporal modeling, spatiotemporal feature fusion, and multi-scale relational learning.

\section{Conclusion}
In this work, we introduced MaGNet, a novel \textbf{Ma}mba dual-hyper\textbf{G}raph \textbf{Net}work designed for stock prediction by synergistically combining advanced temporal modeling with dynamic relational learning.
MaGNet integrates the innovative MAGE block for comprehensive temporal-contextual learning, 2D spatiotemporal attention that enable direct feature-to-feature and stock-to-stock modeling while preserving the intrinsic multivariate structure of financial data, and dual hypergraph framework (TCH and GPH) to dynamically capture temporal-causal relationships and global market structures. Extensive experiments across six major indices validate that MaGNet significantly outperforms state-of-the-art methods in predictive accuracy, investment returns and risk management. 
Future work could explore incorporating alternative data sources and investigating the interpretability of learned hypergraph structures for market insight generation.





\bibliographystyle{ACM-Reference-Format}
\bibliography{sample-base}

\appendix

\section{Experiment Setting Supplement}

\begin{table*}[htb]
\begin{center}
\caption{MaGNet's Hyperparameter Configurations}
\label{tab:Hyperparameters}
    \begin{tabular}{cccccccccc}
\toprule
\textbf{Dataset} &
  \textbf{T} &
  \textbf{\# MAGE} &
  \textbf{\# F. 2D Attn} &
  \textbf{\# TCH} &
  \textbf{\# Hyperedges $M_1$} &
  \textbf{Top-$K$} &
  \textbf{\# S. 2D Attn} &
  \textbf{\# GPH} &
  \textbf{\# Hyperedges $M_2$} \\ \midrule
\textbf{DJIA}       & 20 & 2 & 1 & 1 & 32  & 32  & 1 & 1 & 16 \\
\textbf{HSI}        & 10 & 1 & 1 & 2 & 64  & 64  & 2 & 2 & 32 \\
\textbf{NASDAQ 100} & 10 & 1 & 1 & 2 & 64  & 64  & 1 & 2 & 32 \\
\textbf{S\&P100}    & 30 & 2 & 1 & 2 & 64  & 64  & 1 & 1 & 32 \\
\textbf{CSI300}     & 10 & 1 & 1 & 2 & 64  & 64  & 1 & 1 & 32 \\
\textbf{Nikkei 225} & 10 & 1 & 1 & 1 & 128 & 128 & 2 & 2 & 64 \\ \bottomrule
\end{tabular}
\end{center}
\end{table*}

\subsection{Baseline Descriptions}
\label{Baseline Descriptions}
To evaluate the effectiveness of MaGNet, we compare it against 17 baselines with several state-of-the-art baselines from 3 different categories. These models provide a diverse set of benchmarks to evaluate our method’s performance. 


\textbf{1. Stock Prediction Models (6):}
\begin{itemize}
    \item SFM \cite{zhang2017stock}: State Frequency Memory networks that model price fluctuations across multiple frequencies using frequency-based decomposition.
    \item Adv-ALSTM \cite{feng2018enhancing}: Attentive LSTM with adversarial training for improved robustness against stochastic price movements.
    \item DTML \cite{attention1}: Transformer architecture capturing dynamic inter-stock correlations through multi-level contexts.
    \item ESTIMATE \cite{huynh2023efficient}: Combines wavelet-based hypergraph convolution with memory-enhanced LSTM for non-pairwise stock correlations.
    \item StockMixer \cite{fan2024stockmixer}: MLP-based model that sequentially mixes indicators, temporal patterns, and market correlations.
    \item MASTER \cite{li2024master}: Integrates intra/inter-stock attention with market-guided gating for dynamic correlation capture.
\end{itemize}

\textbf{2. Time Series Models (8):}
\begin{itemize}
    \item GRU \cite{gru}: RNN encoder-decoder with gated recurrent units for sequence-to-vector encoding.
    \item LSTM \cite{gers2000learning}: RNN architecture with gating mechanisms for long-term dependency modeling.
    \item DLinear \cite{zeng2023transformers}: One-layer linear model that directly models temporal relations for long-term forecasting.
    \item TimesNet \cite{wu2022timesnet}: Transforms time series to 2D tensors to model intra/inter-period variations.
    \item PatchTST \cite{nie2022time}: Channel-independent Transformer using patching for improved long-term forecasting.
    \item iTransformer \cite{liu2023itransformer}: Inverted Transformer applying attention across variates rather than time steps.
    \item TimeMixer \cite{wang2024timemixer}: MLP-based model using multiscale mixing to disentangle temporal variations.
    \item TimeXer \cite{wang2024timexer}: Transformer designed for forecasting with exogenous variables using patch-wise and variate-wise attention.
\end{itemize}

\textbf{3. Graph Models (3):}
\begin{itemize}
    \item GCN (Graph Convolutional Network) \cite{kipf2016semi}: Uses first-order spectral graph convolutions for efficient node embedding learning.
    \item GraphSAGE \cite{hamilton2017inductive}: Inductive framework generating embeddings via neighborhood sampling and aggregation.
    \item GAT (Graph Attention Network) \cite{velivckovic2017graph}: Employs masked self-attention to assign weights to neighbors for flexible node embedding.
\end{itemize}

\subsection{Metric Definitions}
\label{Metric Definitions}
\subsubsection{Prediction Metrics}
\begin{equation}
    \text{Accuracy} = \frac{TP + TN}{TP + TN + FP + FN},
\end{equation}
\begin{equation}
    \text{Precision} = \frac{TP}{TP + FP},
\end{equation}
\begin{equation}
    \text{Recall} = \frac{TP}{TP + FN},
\end{equation}
\begin{equation}
    \text{F1} = 2 \cdot \frac{\text{Precision} \cdot \text{Recall}}{\text{Precision} + \text{Recall}} = \frac{2 \cdot TP}{2 \cdot TP + FP + FN},
\end{equation}
\begin{equation}
    \text{AUC} = \int_0^1 \text{TPR}(\text{FPR}) \, d(\text{FPR}),
\end{equation}
where:
\begin{itemize}
    \item TP (True Positives): Correctly predicted positive cases;
    \item TN (True Negatives): Correctly predicted negative cases;
    \item FP (False Positives): Incorrectly predicted as positive;
    \item FN (False Negatives): Incorrectly predicted as negative;
    \item TPR (True Positive Rate) = Recall = $\frac{TP}{TP + FN}$;
    \item FPR (False Positive Rate) = $\frac{FP}{FP + TN}$.
\end{itemize}

\subsubsection{Backtesting Metrics}
\begin{equation}
    \text{Annual Return} = \left[\prod_{t=1}^{T}(1 + r_t)\right]^{\frac{252}{T}} - 1,
\end{equation}
where $r_t$ = return for day t, T = number of trading days, 252 = typical number of trading days per year.

\begin{equation}
    \text{Sharpe Ratio} = \frac{R_p - R_f}{\sigma_p},
\end{equation}
where $R_p$ = annualized portfolio return, $R_f$ = 0.02 (2\% risk-free rate), $\sigma_p$ = annualized standard deviation = $\sigma_{daily} \times \sqrt{252}$.

\begin{equation}
    \text{Calmar Ratio} = \frac{\text{Annual Return}}{|\text{Maximum Drawdown}|},
\end{equation}

\begin{equation}
    \text{Maximum Drawdown} = \min_{t \in [0,T]} \left(\frac{P_t - \max_{s \in [0,t]} P_s}{\max_{s \in [0,t]} P_s}\right),
\end{equation}
where $P_t$ = portfolio value at day t.

\subsection{Implementation Details}
\label{sec:hypersetting}
We passed the final representations $\mathbf{Z}^{\text{GPH}}$ through a FFN followed by softmax to predict the probability distribution of stocks' close price movement directions (rise/fall) for the next trading day. The model was implemented in PyTorch and trained using cross-entropy loss. Hyperparameters were selected via grid search on each dataset, optimizing for validation accuracy. We tuned the number of layers (1–2) for the MAGE blocks, Feature-wise/Stock-wise Spatiotemporal Attention, TCH and GPH. Key settings included embedding dimension $D = 32$, MoE experts=4, spatiotemporal attention channels=4, attention heads=2 (MAGE and CausalMHA) and dropout=0.1. Lookback windows selected from $\{10, 20, 30\}$. Training employed AdamW optimizer with learning rate $1e-4$. We trained for up to 30 epochs using early stopping. Each layer employed residual connections and layer normalization to support stable deep architecture. All baselines used official implementations with default parameters, adapted for stock prediction using cross-entropy loss. Dataset-specific hyperparameter configurations are shown in Table \ref{tab:Hyperparameters}, including Top-$K$ sparsification in TCH; number of hyperedges $M_1$ for TCH and $M_2$ for GPH, respectively.

\subsection{Backtesing Strategy and Configurations }
\label{sec:backtest}

\subsubsection{Dynamic Daily Trading Strategy} 

In this section, we provide a detailed description of the dynamic daily stock trading strategy and hyperparameter configurations used for backtesting.

\begin{algorithm}[htb]
\caption{Dynamic Daily Trading Strategy}
\label{alg:daily-trading-strategy}
\KwData{$N$ stocks, portfolio proportion $p$, stop-loss threshold $q$, conservative ratio $r$, initial capital $1,000,000$, transaction cost rate $\tau = 0.25\%$}

\For{each trading day $t$}{
    \tcp{Prediction Generation}
    $\mathbf{P}_t \gets \text{Model.predict\_probabilities}(\text{all } N \text{ stocks})$\;
    $M \gets |\{s : P_{s,t} > 0.5\}|$\;
    
    \tcp{Portfolio Construction}
    $n_t \gets \begin{cases}
        \lfloor p \times N \rfloor & \text{if } M \geq p \times N \\
        \lfloor r \times M \rfloor & \text{if } p \times N \times q \leq M < p \times N \\
        0 & \text{if } M < p \times N \times q
    \end{cases}$
    
    \tcp{Portfolio Reconstitution \& Rebalancing}
    \eIf{$n_t = 0$}{
        Liquidate all holdings (apply $\tau$)\;
    }{
        $\text{Targets}_t \gets \text{Top-}n_t\text{ stocks by } \mathbf{P}_t$\;
        Liquidate positions $\notin \text{Targets}_t$ (apply $\tau$)\;
        
        \tcp{Equal capital allocation}
        $\text{TargetValue} \gets \frac{\text{TotalPortfolioValue}}{n_t}$\;
        \For{each stock $s \in \text{Targets}_t$}{
            Adjust position of $s$ to $\text{TargetValue}$ (apply $\tau$)\;
        }
    }
}
\end{algorithm}

We initialize each backtest with a capital of $1,000,000$ and apply a transaction cost rate of $0.25\%$ per trade to emulate realistic market frictions. The trading universe consists of all $N$ stocks in each index. The core design is a dynamic daily trading cycle with adaptive portfolio construction and a stop-loss mechanism. Each trading day proceeds through the following steps. The pseudo code of Dynamic Daily Trading Strategy is shown in Algorithm \ref{alg:daily-trading-strategy}. All transaction adjustment incorporate transaction costs.
\begin{itemize}
    \item Prediction Generation: The model outputs the probability of next-day price rise for all stocks, which we use to rank them.
    \item Portfolio Construction with Stop-Loss Mechanism: We define a portfolio selection proportion $p$ (where $0 < p \leq 1$). On each day:
\begin{itemize}
    \item If the number of stocks predicted to rise (with probability $>$ 0.5) is at least $p \times N$, we purchase the top $p \times N$ stocks;
    \item If the number of rising predictions falls into an intermediate zone, specifically $p \times N \times q \leq M < p \times N$ (where $q$ is a stop-loss threshold hyperparameter with $0 < q < 1$), then we adopt a conservative approach: only buy the top $r \times M$ predicted rising stocks (with $0 \leq r \leq 1$);
    \item If the number of rising predictions is below $p \times N \times q$, we do not buy new positions that day and liquidate all current holdings to avoid downside exposure.
\end{itemize}
    \item Portfolio Reconstitution: Positions excluded from the new targets are liquidated with proceeds credited to cash. New target stocks are then purchased with equal capital allocation, subject to current available cash.
    \item Portfolio Rebalancing: To maintain equal-capital allocations, we adjusts positions daily—selling excess holdings that exceed target allocation and purchasing additional shares for under-allocated positions.
\end{itemize}
This backtesting framework enables direct comparison of model predictions in a realistic trading environment, providing a robust evaluation of each model’s profitability performance under real-world conditions. 

\begin{table}[htb]
\centering
\caption{MaGNet'S Backtesting Hyperparameters}
\label{tab:backtest hyperparameters}
\begin{tabular}{cccc}
\toprule
\textbf{Dataset}    & \textbf{p} & \textbf{q} & \textbf{r} \\ \midrule
\textbf{DJIA}       & 1          & 0.05       & 0          \\
\textbf{HSI}        & 1          & 0.05       & 0          \\
\textbf{NASDAQ 100} & 1          & 0.4        & 1          \\
\textbf{S\&P 100}    & 1          & 0.05       & 0          \\
\textbf{CSI 300}     & 1          & 0.05       & 0          \\
\textbf{Nikkei 225} & 1          & 0.7        & 1         \\ \bottomrule
\end{tabular}
\end{table}

\subsubsection{Backtesting Hyperparameter Configurations}

For each (model, dataset) pair, we perform a grid search on the validation set over three parameters in dynamic trading strategy:
\begin{itemize}
    \item Portfolio Selection Ratio $p \in \{0.05, 0.10, \ldots, 1.0\}$;
    \item Stop-loss Threshold $q \in \{0.05, 0.10, \ldots, 0.95\}$;
    \item Rising Ratio for Partial Entry $r \in \{0.0, 0.05, \ldots, 1.0\}$.
\end{itemize}
The combination yielding the highest Sharpe ratio on the validation set is selected and applied to the test set for final evaluation. The backtesting hyperparameters for MaGNet on each dataset are shown in Table \ref{tab:backtest hyperparameters}.

\section{Additional Experimental Results}
\label{Additional Experimental Results}
This section presents the additional prediction and backtesting results on the S\&P 100, CSI 300, and Nikkei 225 indices, and results of ablation studies on all six indices.

\subsection{Additional Prediction Results}
\label{Additional Prediction Results}

\begin{table*}[htb]
\begin{center}
\caption{Prediction performance comparison on S\&P 100, CSI 300, and Nikkei 225. The best results are in bold and the second-best results are underlined.}
\label{tab:Prediction results 2}
\setlength{\tabcolsep}{1mm}
\begin{tabular}{cccccccccccccccc}
\toprule
\multicolumn{1}{c}{\multirow{2}{*}{Model}} & \multicolumn{5}{c}{S\&P 100}                                                       & \multicolumn{4}{c}{CSI 300}                                       &                & \multicolumn{5}{c}{Nikkei 225}                                                      \\ \cmidrule(lr){2-6} \cmidrule(lr){7-11} \cmidrule(lr){12-16}
\multicolumn{1}{c}{}                       & ACC            & PRE            & REC            & F1             & AUC            & ACC            & PRE            & REC            & F1             & AUC            & ACC            & PRE            & REC             & F1             & AUC            \\ \midrule
GRU                                        & 51.57          & 52.75          & 76.71          & 62.52          & 50.46          & 53.00          & 52.85          & 10.86          & 18.02          & 52.26          & 51.28          & 52.85          & 48.61           & 50.64          & 52.11          \\
LSTM                                       & 52.37          & 52.76          & 91.07          & 66.81          & 50.76          & 53.06          & 53.17          & 10.79          & 17.94          & 52.88          & 50.12          & 52.04          & 38.14           & 44.02          & 50.92          \\
DLinear                                    & 52.29          & 52.61          & 94.12          & 67.50          & {\ul 51.21}    & 52.97          & 51.28          & 22.25          & 31.04          & 52.43          & 51.18          & 51.36          & 95.00           & 66.68          & 49.94          \\
TimesNet                                   & 52.26          & 52.87          & 85.72          & 65.40          & 50.42          & 53.02          & 50.91          & 33.94          & 40.73          & 53.22          & {\ul 51.66}    & 52.82          & 56.13           & 54.43          & 52.46          \\
PatchTST                                   & 50.78          & 52.64          & 64.71          & 58.05          & 49.93          & 51.86          & 49.18          & 37.29          & 42.42          & 51.77          & 49.31          & 50.66          & 54.29           & 52.41          & 48.94          \\
iTransformer                                    & 52.32          & 52.65          & 93.56          & 67.38          & 50.04          & 52.27          & 49.65          & 26.46          & 34.52          & 51.95          & 49.09          & 50.55          & 45.69           & 48.00          & 48.92          \\
TimeMixer                                     & 52.54          & 52.69          & 96.42          & 68.14          & 50.28          & 52.78          & 52.19          & 8.42           & 14.50          & 52.99          & 51.42          & 51.42          & \textbf{100.00} & \textbf{67.92} & 48.38          \\
TimeXer                                     & 52.48          & 52.66          & 96.24          & 68.07          & 50.27          & 52.98          & 52.52          & 11.58          & 18.97          & 52.77          & 51.42          & 51.42          & \textbf{100.00} & \textbf{67.92} & 51.15          \\
GCN                                        & 50.31          & 52.03          & 71.81          & 60.34          & 50.80          & {\ul 54.13}    & {\ul 53.50}    & 27.04          & 35.92          & 52.33          & 50.97          & 52.64          & 46.18           & 49.20          & 51.26          \\
GraphSAGE                                     & 52.16          & 52.66          & 90.29          & 66.52          & 49.78          & 53.06          & 52.24          & 14.96          & 23.26          & 52.05          & 51.33          & {\ul 52.88}    & 49.05           & 50.89          & 52.91          \\
GAT                                        & 49.77          & 52.41          & 49.71          & 51.02          & 50.09          & 50.64          & 47.90          & \textbf{43.46} & \textbf{45.57} & 51.51          & 51.03          & 52.12          & 58.48           & 55.12          & {\ul 53.03}    \\
SFM                                        & 50.76          & \textbf{53.84} & 52.50          & 53.16          & 50.95          & 53.12          & 50.78          & {\ul 39.74}    & {\ul 44.59}    & {\ul 53.95}    & 50.72          & 52.27          & 45.23           & 48.50          & 51.05          \\
Adv-ALSTM                                     & 52.45          & 52.65          & 94.91          & 67.71          & 50.01          & 53.00          & 51.82          & 17.49          & 26.05          & 52.89          & 51.41          & 52.59          & 56.55           & 54.44          & 51.81          \\
DTML                                       & 52.44          & 52.88          & 88.68          & 66.25          & 51.03          & 53.38          & 52.74          & 18.83          & 27.75          & 53.59          & 51.05          & 52.19          & 57.10           & 54.54          & 51.51          \\
ESTIMATE                                     & {\ul 52.59}    & 52.71          & {\ul 96.70}    & {\ul 68.23}    & 50.25          & 52.37          & 46.64          & 1.11           & 2.16           & 49.40          & 50.22          & 51.83          & 45.23           & 48.31          & 50.43          \\
StockMixer                                    & 52.53          & 52.69          & 95.99          & 68.04          & 50.98          & 52.77          & 50.52          & 32.93          & 39.87          & 53.24          & 51.40          & 51.47          & 96.11           & 67.04          & 50.47          \\
MASTER                                      & 51.66          & {\ul 53.72}    & 58.97          & 56.22          & \textbf{51.68} & 52.94          & 51.05          & 25.07          & 33.63          & 53.13          & 51.60          & 51.56          & 97.20           & 67.38          & 52.47          \\
MaGNet                                     & \textbf{53.14} & 53.00          & \textbf{97.00} & \textbf{68.55} & 49.36          & \textbf{54.90} & \textbf{54.03} & 34.58          & 42.17          & \textbf{54.59} & \textbf{54.02} & \textbf{54.96} & 58.59           & 56.72          & \textbf{53.55}\\ \bottomrule
\end{tabular}
\end{center}
\end{table*}

Experiment results on S\&P 100, CSI 300, and Nikkei 225 indices are shown in Table \ref{tab:Prediction results 2}, which further validate the generalizability and profitability of MaGNet. It achieves the highest accuracy on CSI 300 (54.90\%), HSI (54.19\%), and Nikkei 225 (54.02\%). MaGNet also shows strong recall on S\&P 100 (97.00\%) and balanced precision–recall trade-offs across datasets. These results confirm that the integration of MAGE’s comprehensive temporal modeling with the dual-hypergraph framework effectively captures both localized temporal–causal dependencies and global market structures, yielding robust and generalizable stock movement predictions across diverse markets, with especially strong improvements on larger universes (CSI 300, Nikkei 225).

\subsection{Additional Backtesting Results}
\label{Additional Backtesting Results}

\begin{table*}[htb]
\centering
\caption{Backtesting performance comparison on S\&P 100, CSI300, and Nikkei 225. The best results are in bold and the second-best results are underlined.}
\label{tab:Backtest results 2}
\begin{tabular}{ccccccccccccc}
\toprule
\multicolumn{1}{c}{\multirow{2}{*}{Model}} & \multicolumn{4}{c}{S\&P 100}                                   & \multicolumn{4}{c}{CSI 300}                                    & \multicolumn{4}{c}{Nikkei 225}                                     \\ \cmidrule(lr){2-5} \cmidrule(lr){6-9} \cmidrule(lr){10-13}
\multicolumn{1}{c}{}                       & AR             & SR            & CR            & MDD           & AR             & SR            & CR            & MDD           & AR             & SR            & CR            & MDD           \\ \midrule
GRU                                        & 13.59          & 1.12          & 1.98          & 6.85          & 16.35          & {\ul 1.27}    & 1.83          & 8.93          & 9.97           & 0.52          & 0.92          & 10.80         \\
LSTM                                       & 13.70          & 1.06          & 2.03          & 6.74          & 20.22          & 1.26          & 2.79          & 7.25          & 14.23          & 0.47          & 0.66          & 21.61         \\
DLinear                                    & 10.74          & 0.80          & 1.84          & 5.84          & 16.61          & 1.15          & 1.97          & 8.43          & 7.13           & 0.22          & 0.31          & 22.91         \\
TimesNet                                   & 7.00           & 0.45          & 1.06          & 6.58          & 7.50           & 0.55          & 0.77          & 9.74          & 3.40           & 0.07          & 0.21          & 15.94         \\
PatchTST                                   & 1.49           & -0.15         & 0.83          & {\ul 1.80}    & 0.81           & -0.87         & 1.04          & \textbf{0.78} & 3.18           & 0.10          & 0.49          & 6.47          \\
iTransformer                                    & 11.81          & 0.90          & 2.15          & 5.50          & 12.74          & 0.74          & 2.21          & 5.77          & 5.17           & 0.32          & 1.28          & \textbf{4.04} \\
TimeMixer                                     & 14.87          & 1.19          & 2.75          & 5.40          & {\ul 21.38}    & 0.90          & 2.24          & 9.55          & 15.18          & 0.57          & 0.67          & 22.57         \\
TimeXer                                     & 14.27          & 1.13          & 2.55          & 5.60          & 9.03           & 0.61          & 1.37          & 6.60          & 15.18          & 0.57          & 0.67          & 22.57         \\
GCN                                        & {\ul 16.99}    & 0.91          & 1.58          & 10.79         & 14.55          & 1.18          & 2.27          & 6.41          & 15.33          & {\ul 0.81}    & 1.04          & 14.78         \\
GraphSAGE                                     & 10.24          & 0.79          & 1.84          & 5.56          & 12.47          & 0.77          & 1.33          & 9.35          & 13.96          & 0.56          & 0.66          & 21.05         \\
GAT                                        & 9.87           & 0.69          & 1.44          & 6.84          & 9.89           & 0.44          & 0.60          & 16.40         & 16.39          & 0.57          & 0.67          & 24.42         \\
SFM                                        & 1.98           & -0.01         & 1.45          & \textbf{1.36} & 14.61          & 0.98          & 2.09          & 7.00          & 6.43           & 0.76          & {\ul 1.40}    & {\ul 4.59}    \\
Adv-ALSTM                                     & 15.65          & {\ul 1.27}    & {\ul 2.97}    & 5.27          & 16.28          & 0.84          & 1.13          & 14.35         & 5.69           & 0.20          & 0.32          & 17.57         \\
DTML                                       & 12.69          & 0.97          & 2.12          & 5.98          & 14.00          & 1.04          & {\ul 2.47}    & 5.67          & 6.21           & 0.33          & 0.60          & 10.37         \\
ESTIMATE                                     & 15.58          & 1.16          & 2.93          & 5.33          & 15.37          & 0.59          & 1.58          & 9.76          & {\ul 19.57}    & 0.66          & 0.75          & 26.17         \\
StockMixer                                    & 11.83          & 0.92          & 2.09          & 5.66          & 12.59          & 1.11          & 2.11          & 5.97          & 9.32           & 0.32          & 0.41          & 22.95         \\
MASTER                                      & 4.10           & 0.35          & 1.08          & 3.78          & 15.65          & 1.15          & 2.39          & 6.54          & 17.58          & 0.69          & 0.78          & 22.58         \\
MaGNet                                     & \textbf{17.14} & \textbf{1.40} & \textbf{3.27} & 5.24          & \textbf{22.60} & \textbf{1.32} & \textbf{4.36} & {\ul 5.19}    & \textbf{19.58} & \textbf{1.14} & \textbf{2.10} & 9.31       \\ \bottomrule  
\end{tabular}
\end{table*}

Backtesting results om S\&P 100, CSI300, and Nikkei 225 are reported in Table \ref{tab:Backtest results 2}. On these indices, MaGNet attains the highest Sharpe Ratios of 1.40 (S\&P 100), 1.32 (CSI300), and 1.14 (Nikkei 225). It also records the largest Annual Returns among the compared methods on S\&P 100 (17.14\%), CSI300 (22.60\%), and Nikkei 225 (19.58\%). 
Furthermore, MaGNet achieves the highest Calmar ratio across these indices, indicating superior returns with moderate drawdowns (5.19\% on CSI 300).
These results reinforce the main-text findings that the MAGE block's temporal modeling combined with the dual-hypergraph architecture yields consistently higher risk-adjusted returns and competitive downside control, particularly on larger market universes.

\subsection{Complete Results of Ablation Studies}
\label{Complete Results of Ablation Studies}

\begin{table*}[htb]
\begin{center}
\caption{Complete ablation results. The best results are in bold and the second-best results are underlined.}
\label{tab:Complete Ablation Results}
\begin{tabular}{cllllllllll}
\toprule
\multirow{2}{*}{Dataset}    & \multicolumn{1}{c}{\multirow{2}{*}{Component}} & \multicolumn{5}{c}{Prediction}                                                      & \multicolumn{4}{c}{Backtesting}                                \\ \cmidrule(lr){3-7} \cmidrule(lr){8-11}
                            & \multicolumn{1}{c}{}                           & ACC            & PRE            & REC             & F1             & AUC            & AR             & SR            & CR            & MDD           \\ \midrule
\multirow{5}{*}{DJIA}       & w/o MAGE                                       & 52.73          & 53.13          & 94.13           & 67.92          & 49.61          & 9.38           & 0.70          & 1.34          & 7.02          \\
                            & w/o F. 2D Attn                                 & {\ul 53.10}    & {\ul 53.15}    & {\ul 99.30}     & {\ul 69.24}    & {\ul 50.84}    & 15.61          & 1.30          & 2.83          & 5.51          \\
                            & w/o TCH                                        & 53.03          & 53.12          & 99.19           & 69.19          & 49.68          & {\ul 18.32}    & {\ul 1.54}    & {\ul 3.62}    & \textbf{5.07} \\
                            & w/o GPH                                        & 52.96          & 53.08          & 99.19           & 69.16          & 48.73          & 18.16          & 1.53          & 3.38          & 5.37          \\
                            & MaGNet                                         & \textbf{53.16} & \textbf{53.16} & \textbf{100.00} & \textbf{69.42} & \textbf{51.10} & \textbf{19.92} & \textbf{1.70} & \textbf{3.93} & \textbf{5.07} \\ \hline
\multirow{5}{*}{HSI}        & w/o MAGE                                       & 52.43          & 51.15          & 24.01           & 32.68          & 50.00          & 6.98           & 0.35          & 0.49          & 14.23         \\
                            & w/o F. 2D Attn                                 & 52.34          & 50.80          & {\ul 28.09}     & {\ul 36.18}    & 50.95          & 6.44           & 0.31          & {\ul 0.67}    & {\ul 9.61}    \\
                            & w/o TCH                                        & 52.53          & 52.30          & 14.49           & 22.69          & 50.44          & {\ul 8.59}     & 0.29          & 0.39          & 21.80         \\
                            & w/o GPH                                        & {\ul 53.76}    & \textbf{54.07} & 25.42           & 34.59          & {\ul 53.60}    & 8.26           & {\ul 0.36}    & 0.59          & 13.89         \\
                            & MaGNet                                         & \textbf{54.19} & {\ul 52.85}    & \textbf{43.90}  & \textbf{47.96} & \textbf{54.12} & \textbf{12.25} & \textbf{0.66} & \textbf{1.33} & \textbf{9.20} \\ \hline
\multirow{5}{*}{NASDAQ 100} & w/o MAGE                                       & {\ul 52.97}    & {\ul 52.90}    & 88.98           & 66.36          & 50.29          & 9.68           & 0.56          & 1.09          & {\ul 8.85}    \\
                            & w/o F. 2D Attn                                 & 52.17          & 52.26          & 95.11           & 67.46          & 48.46          & {\ul 13.41}    & {\ul 0.80}    & {\ul 1.35}    & 9.96          \\
                            & w/o TCH                                        & 52.73          & 52.56          & {\ul 95.36}     & {\ul 67.77}    & {\ul 50.97}    & 9.49           & 0.50          & 0.92          & 10.31         \\
                            & w/o GPH                                        & 52.65          & 52.61          & 92.16           & 66.99          & 47.98          & 8.44           & 0.48          & 0.78          & 10.76         \\
                            & MaGNet                                         & \textbf{53.72} & \textbf{53.09} & \textbf{96.14}  & \textbf{68.41} & \textbf{52.24} & \textbf{17.09} & \textbf{1.05} & \textbf{2.09} & \textbf{8.18} \\ \hline
\multirow{5}{*}{S \&P 100}  & w/o MAGE                                       & 52.74          & 52.89          & 93.63           & 67.59          & \textbf{50.14} & 10.63          & 0.81          & 1.93          & 5.52          \\
                            & w/o F. 2D Attn                                 & {\ul 52.97}    & {\ul 52.95}    & 95.84           & {\ul 68.21}    & 49.23          & {\ul 13.61}    & {\ul 1.13}    & {\ul 2.59}    & \textbf{5.24} \\
                            & w/o TCH                                        & 52.10          & 52.46          & 95.88           & 67.82          & 48.81          & 12.97          & 1.01          & 2.08          & 6.23          \\
                            & w/o GPH                                        & 52.42          & 52.61          & {\ul 96.84}     & 68.18          & 48.98          & 12.49          & 0.97          & 2.38          & \textbf{5.24} \\
                            & MaGNet                                         & \textbf{53.14} & \textbf{53.00} & \textbf{97.00}  & \textbf{68.55} & {\ul 49.36}    & \textbf{17.14} & \textbf{1.40} & \textbf{3.27} & \textbf{5.24} \\ \hline
\multirow{5}{*}{CSI 300}    & w/o MAGE                                       & {\ul 54.40}    & \textbf{60.46} & 11.90           & 19.89          & 51.40          & 8.10           & 0.57          & 1.48          & {\ul 5.45}    \\
                            & w/o F. 2D Attn                                 & 53.07          & 52.29          & 14.89           & 23.17          & 50.93          & {\ul 18.36}    & {\ul 0.94}    & 1.48          & 12.42         \\
                            & w/o TCH                                        & 52.91          & 51.04          & 23.83           & 32.49          & {\ul 52.68}    & 12.12          & 0.66          & {\ul 2.06}    & 5.88          \\
                            & w/o GPH                                        & 53.25          & 51.43          & {\ul 30.26}     & {\ul 38.10}    & 51.67          & 11.84          & 0.62          & 1.13          & 10.47         \\
                            & MaGNet                                         & \textbf{54.90} & {\ul 54.03}    & \textbf{34.58}  & \textbf{42.17} & \textbf{54.59} & \textbf{22.60} & \textbf{1.32} & \textbf{4.36} & \textbf{5.19} \\ \hline
\multirow{5}{*}{Nikkei 225} & w/o MAGE                                       & 52.74          & {\ul 54.60}    & 48.10           & 51.14          & {\ul 53.38}    & 3.26           & 0.07          & 0.18          & 18.56         \\
                            & w/o F. 2D Attn                                 & 53.05          & 54.47          & 53.00           & 53.73          & 52.69          & {\ul 14.25}    & {\ul 0.59}    & 0.63          & 22.57         \\
                            & w/o TCH                                        & 51.01          & 52.34          & 52.77           & 52.56          & 50.98          & 9.24           & 0.55          & {\ul 1.05}    & \textbf{8.84} \\
                            & w/o GPH                                        & {\ul 53.13}    & 54.15          & {\ul 57.67}     & {\ul 55.86}    & 51.97          & 2.14           & 0.01          & 0.09          & 22.94         \\
                            & MaGNet                                         & \textbf{54.02} & \textbf{54.96} & \textbf{58.59}  & \textbf{56.72} & \textbf{53.55} & \textbf{19.58} & \textbf{1.14} & \textbf{2.10} & {\ul 9.31}   \\ \bottomrule
\end{tabular}
\end{center}
\end{table*}

Comprehensive ablation results on all six indices are reported in Table \ref{tab:Complete Ablation Results}. The results reveal that each component contributes critically to MaGNet's performance, with varying impacts across different markets. 

Removing the MAGE block causes the most substantial degradation, particularly in trading performance—Annual Return on DJIA drops from 19.92\% to 9.38\% and Sharpe Ratio from 1.70 to 0.70, while recall on CSI 300 plummets from 34.58\% to 11.90\%, confirming MAGE block's essential role in capturing complex temporal dynamics for both prediction and profitability.

The Feature-wise 2D Spatiotemporal Attention proves crucial for maintaining stable trading performance, as Annual Return on S\&P 100 drops from 17.14\% to 13.61\% without it, validating its effectiveness in preserving cross-feature spatiotemporal structures.

The dual-hypergraph components exhibit complementary and market-specific contributions: TCH removal significantly impacts markets with strong temporal dependencies (AR on HSI drops from 12.25\% to 8.59\%), while GPH removal substantially degrades performance in markets with prominent global patterns (AR on NASDAQ 100 falls from 17.09\% to 8.44\%).

Overall, the ablation studies demonstrate that MaGNet’s superior performance stems from the synergistic integration of advanced temporal modeling through MAGE block, 2D spatiotemporal feature fusion, and multi-scale relational learning via the dual-hypergraph framework, with each module addressing distinct aspects of the complex stock prediction challenge.

\end{document}